%% file: iclr2026_conference.tex
\documentclass{article} 
\usepackage{iclr2026_conference,times}

\input{math_commands.tex}

\usepackage{graphicx}
\usepackage{hyperref}
\usepackage{url}
\usepackage{booktabs}
\usepackage{tabularx}
\usepackage{multirow}

\title{Just A Rather Very Intelligent Spoken Agent}

\author{
Chen Chen \quad Zhehuai Chen \\
\ NVIDIA
}

\iclrfinalcopy 
\begin{document}

\maketitle

\begin{abstract}

Long-horizon AI agents are becoming increasingly capable, yet their interaction with users remains surprisingly thin. In most workflows, users give an initial instruction, receive only selective textual updates,  and lose a clear sense of what the agent is doing or when to step in. This leaves a missing part in the current agent ecosystem: an always-on Jarvis-style mediator that keeps the agent continuously reachable to the user. Such a mediator should support real-time spoken interaction with the user, answer questions without interrupting the worker, proactively report progress or confusion, and inject user guidance back into the agent's execution when useful. In this work, we introduce \emph{JarvisBench}, a benchmark for measuring the dual value of mediation in long-horizon agent workflows. JarvisBench contains two complementary tracks: an agent-collaboration track that measures whether mediation improves downstream task completion, and a user-interaction track that measures whether mediation makes ongoing execution more understandable, responsive, and accessible to users. We instantiate the benchmark with a modular reference Jarvis prototype and evaluate it on 34 text-only WildClaw tasks executed in OpenClaw. Preliminary results with GPT-5.5, Claude Opus 4.7, Gemini-based, and GPT-based worker agents suggest that Jarvis-style mediation can provide trace-grounded responses to user questions and improve task performance when sparse user guidance is injected at appropriate moments. The results also show that effectiveness depends strongly on the mediator's LLM brain, highlighting both the promise of this missing middle layer and the need for broader community effort. Demo page \url{https://cchen1436.github.io/jarvis}

\end{abstract}

\begin{figure*}[h!]
\centering
\includegraphics[width=14cm]{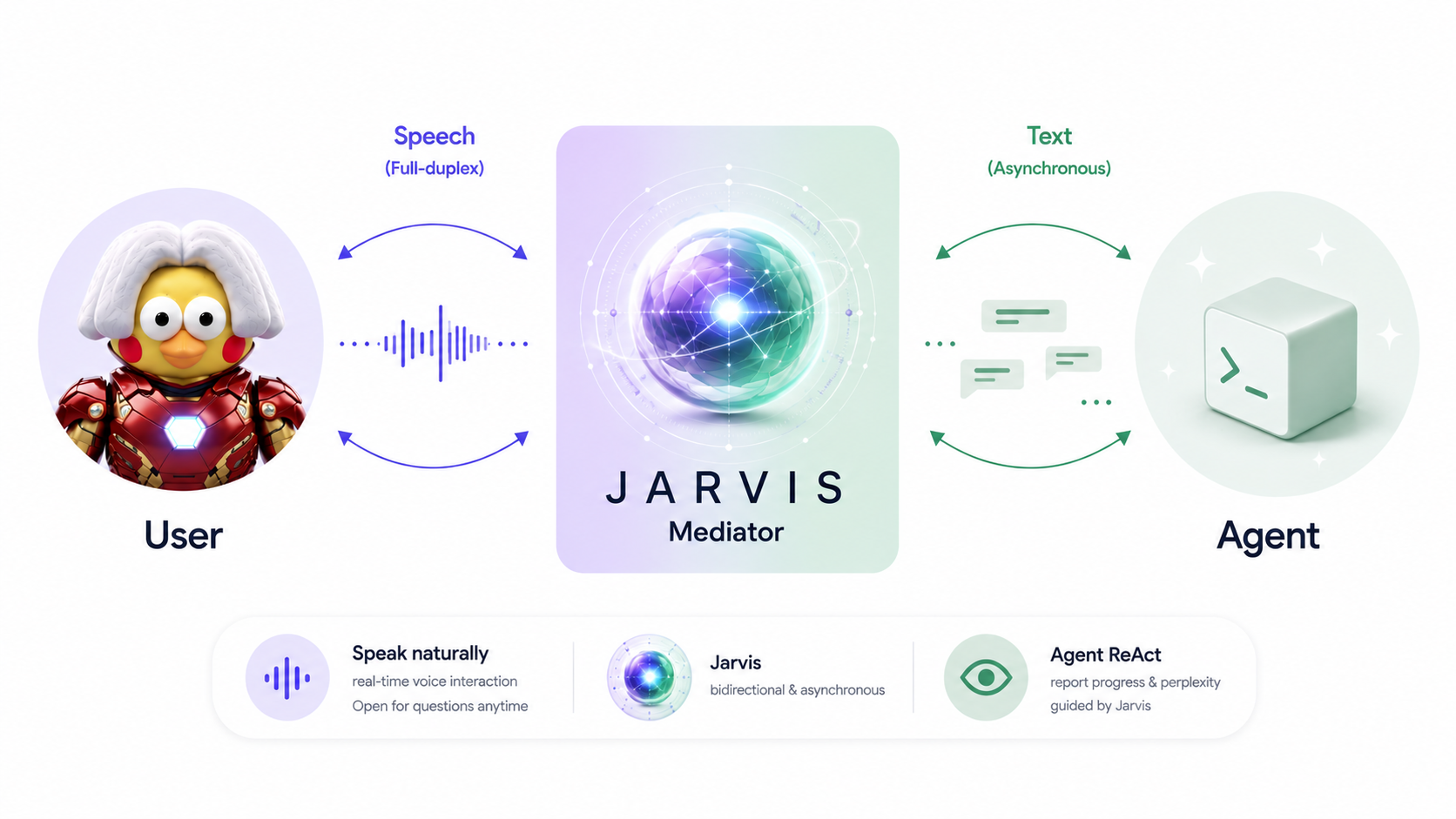}

\label{fig:fd-latent}
\vspace{-5pt}
\end{figure*}

\section{Introduction}

Recent advances in AI agents have rapidly expanded their capabilities across diverse domains, from solving complex programming tasks to assisting users in everyday activities~\citep{huang2024understanding}. Despite their growing competence and deployment potential, their interaction paradigm remains largely unchanged. Most existing agents operate in a turn-based manner: users issue instructions, agents work autonomously for extended periods, and communication occurs only through sparse textual updates, if any. Such delayed and intermittent interaction often leaves users out of the loop, making it difficult to understand agent progress, correct errors early, or steer the execution before the agent drifts away from the user's intent. \par

Science fiction has long imagined a different paradigm for human-agent interaction. In Iron Man, J.A.R.V.I.S. functions as an always-on intelligent collaborator that communicates with user in a spoken manner: proactively sharing progress, answering questions anytime, and executing commands through natural voice interaction. This hands-free, real-time experience enables fantastic collaboration without pausing their ongoing activities. In contrast, today's agents remain largely turn-based and screen-bound: users can launch an agent, but cannot easily stay connected to it. \par

We argue that this contrast points to a missing interface in current long-horizon agent workflows. The issue is not simply that agents need stronger models or larger agent teams. Rather, users and agents need a continuous communication layer between them. On the user side, this layer should make agent execution reachable through natural speech: users should be able to ask what is happening, request clarification, and receive progress reports without inspecting the full execution trace. On the agent side, this layer should monitor the worker agent's working trajectory, surface confusion or failure to the user when guidance may help, and carry concise user feedback back into the worker's ongoing ReAct timing. \par

In this work, we introduce \emph{JarvisBench}, a benchmark for measuring the value of a Jarvis-style mediator between the user and a long-horizon worker agent. JarvisBench is organized into two tracks. The \emph{user-interaction track} evaluates whether the mediator can provide timely, trace-grounded, and task-consistent responses to users during execution. The \emph{agent-collaboration track} evaluates whether the mediator can help the worker agent recover from failure-prone states by consulting the user and injecting concise guidance back into the ongoing process. This two-track design makes it possible to test different LLM brains, speech modules, worker agents, monitoring policies, or memory designs under the same mediation role. To instantiate the benchmark, we build a modular reference Jarvis prototype. The user-facing side supports spoken interaction, while the agent-facing side receives textual ReAct-style updates from the worker agent, maintains execution context, and opens interaction checkpoints when user questions or expert guidance should be handled. This prototype is not intended to be the only possible implementation of Jarvis; it is a reference system for running the benchmark and exposing the required logs. \par

We instantiate JarvisBench on 34 text-only WildClaw~\citep{wildclaw} cases executed in OpenClaw or Hermes platforms, covering practical long-horizon tasks that require planning, tool use, recovery, and safety-aware decisions. With simulated nonexpert users, Jarvis provides useful progress and topic-related responses during execution, suggesting that mediation can make agent work more transparent without requiring users to inspect the full trajectory. With simulated expert users, Jarvis also improves task outcomes by reporting failure-prone states and relaying concise user guidance back to the worker agent. The effect depends strongly on the capability of the Jarvis brain: with a GPT-level mediator, the Overall Score of a Claude Opus 4.7-based worker agent increases from 64.01 to 75.79 on WildClaw. Case analysis further suggests that useful guidance does not need to be highly specialized; even when the expert user cannot access the full ReAct trajectory, simple suggestions based on Jarvis's reports can help the worker agent escape local failure modes.

\section{JarvisBench: A Benchmark for User-Agent Mediation}

We formulate Jarvis-style mediation as a plug-and-play benchmark rather than a fixed agent architecture. JarvisBench evaluates a mediator placed between a user and a long-horizon worker agent. The mediator may observe the worker agent's execution trace, communicate with the user through speech, and optionally inject user guidance back into the worker's ongoing context. A system tested under JarvisBench can replace the mediator's LLM brain, speech interface, monitoring policy, memory design, or worker agent, as long as it exposes the logs required by the evaluation protocol.

The benchmark is designed around two questions. First, does the mediator help the worker agent complete tasks more successfully? Second, does the mediator make the agent's ongoing execution more understandable and accessible to the user? We therefore organize JarvisBench into two tracks: an \emph{agent-collaboration track} and a \emph{user-interaction track}. The first track measures the value of mediation for task completion; the second measures the value of mediation for real-time user experience.

\subsection{Track 1: Agent-Collaboration}

The agent-collaboration track evaluates whether a mediator improves downstream task completion. The same worker agent is evaluated in two settings: the original OpenClaw execution loop, and the same loop augmented with Jarvis mediation. During execution, Jarvis monitors the worker agent's ReAct trajectory and identifies sparse failure-sensitive checkpoints, such as repeated tool failures, stalled progress, ambiguous observations, risky irreversible actions, or low-confidence finalization. When such a checkpoint is triggered, Jarvis summarizes the relevant execution state, asks the expert user for concise guidance, and injects the feedback back into the worker agent's context.

This track is deliberately different from evaluating a multi-agent system. Jarvis is not allowed to solve the task independently, debate with the worker agent, or replace the worker's trajectory with its own plan. Its role is to expose the worker's difficulty to the user and carry user feedback back into the ongoing execution. We report the following metrics:
\begin{itemize}
    \item \textbf{Overall Score ($\%$)}: the WildClaw task-completion score of the worker agent, judged under the original WildClaw evaluation protocol.
    \item \textbf{Relative gain ($\Delta$)}: the absolute improvement over the same worker agent without Jarvis mediation.
    \item \textbf{\# Full-score}: the number of tasks receiving the maximum task score.
    \item \textbf{\# Interv.}: the number of user-guidance interventions injected by Jarvis during execution.
\end{itemize}
Table~\ref{tab:wildclaw-jarvis-results} reports this track.

\subsection{Track 2: User-Interaction}

The user-interaction track evaluates whether a mediator can provide useful real-time interaction to a nonexpert user while the worker agent continues solving the task. To isolate the mediator's user-facing ability, we fix the worker agent and vary the Jarvis brain or other mediator components. The nonexpert user knows the original request but does not provide task-solving guidance. Instead, the user asks short natural questions during execution, alternating between progress-oriented questions and topic-oriented questions.

Progress questions test whether Jarvis can explain what the worker agent is currently doing based on the visible execution trace. Topic questions test whether Jarvis can answer task-relevant background or context questions based on the original user request and available task context. For each answer, we save the evidence available to Jarvis at the time of response and use separate LLM-judge prompts for progress and topic evaluation. We report:
\begin{itemize}
    \item \textbf{Progress}: the average 1--5 score for whether the answer is grounded in the current execution trace.
    \item \textbf{Prog. $\geq$4}: the fraction of progress answers rated at least 4.
    \item \textbf{Topic}: the average 1--5 score for whether the answer is consistent with the original task request and task context.
    \item \textbf{Topic $\geq$4}: the fraction of topic answers rated at least 4.
    \item \textbf{Latency}: the elapsed time from the end of the user's spoken question to the start of Jarvis's spoken reply.
\end{itemize}
Table~\ref{tab:nonexpert-user-eval} reports this track.

\subsection{Plug-and-Play Evaluation Protocol}

JarvisBench supports component-level replacement. To test a new LLM as the Jarvis brain, one keeps the worker agent, user simulator, checkpoint schedule, judge prompts, and speech stack fixed, and replaces only the mediator LLM. To test a new worker agent, one runs the same task set with and without a fixed Jarvis mediator. To test a new speech or conversation module, one keeps the Jarvis brain fixed and reports user-interaction quality together with latency. To test a new monitoring or intervention policy, one keeps the worker agent and user simulator fixed, and measures whether the policy improves task completion without introducing unnecessary interventions.

This protocol turns Jarvis from a single proposed system into a benchmarkable interface for user-agent mediation. Different systems may instantiate the mediator with a cascaded speech pipeline, an omni model, a stronger text LLM, a different memory design, or a new checkpointing policy. As long as they follow the same two-track protocol and expose the required interaction logs, their gains can be compared along the two axes that matter most: whether the user is better connected to the execution, and whether the worker agent completes tasks more successfully.
\section{Jarvis Prototype}

We build a modular reference prototype of Jarvis. This prototype is not intended to define a single required architecture for user-agent mediation. Instead, it provides one practical implementation that exposes the logs, checkpoints, and interaction channels needed by the two benchmark tracks. Each module can be replaced independently, including the Jarvis brain, speech interface, monitoring policy, or worker agent, while keeping the evaluation protocol unchanged.

\subsection{Cascaded Full-Duplex Speech Interface}

We implement the user-facing interface as a cascaded speech pipeline. A streaming ASR module converts the user's speech into partial transcripts, and a semantic voice activity detection module decides whether the user has finished speaking, intends to interrupt, or is still holding the floor. In our prototype, we use SoulX-Duplug~\citep{soul} as the semantic VAD module and Chatterbox~\citep{chatterboxtts2025} for streaming text-to-speech generation. Once the semantic VAD detects a completed user utterance, the Jarvis brain receives the transcript and generates a response that is spoken back to the user.

This design allows Jarvis to stay in an always-on listening state during agent execution. The user can ask questions without manually opening the agent trace, and Jarvis can answer without stopping the worker agent unless the benchmark checkpoint requires a pause. The same interface also supports proactive speech: when Jarvis detects a failure-sensitive state in the worker trajectory, it can initiate a short spoken interaction with the user. In the user-interaction track, latency is measured from the end of the user's spoken question to the start of Jarvis's spoken reply. Although we use a cascaded pipeline for reliability and controllability, the speech interface is replaceable; future submissions may use an omni model or a different ASR/VAD/TTS stack under the same evaluation protocol.

\subsection{Trace Monitoring and Context Management}

The agent-facing side of Jarvis receives textual execution updates from the worker agent, including ReAct-style reasoning summaries, tool calls, observations, errors, and proposed next actions. The Jarvis brain uses this stream for two purposes. First, it answers user questions about the worker's progress or the task context. Second, it detects moments where user guidance may help the worker recover from a failure-prone trajectory. Jarvis does not execute tools or solve the task independently; its output to the worker is limited to concise user-grounded guidance at selected checkpoints.

To keep this process stable, the prototype separates dialogue context from execution context. The dialogue context stores recent user-Jarvis exchanges and user preferences. The execution context stores the worker's recent actions, observations, failures, and current plan. A small control state records whether Jarvis is answering a user question, waiting for expert guidance, or preparing an injected instruction for the worker. This separation is important because user-facing dialogue and agent-facing monitoring impose different context demands: the former should be natural and concise, while the latter should remain grounded in the execution trace.

\subsection{Interaction Checkpointing}

Real-time interaction introduces latency, especially when the user is simulated by an LLM and the response is spoken through TTS. To make evaluation reproducible, we implement interaction checkpointing as a cooperative pause mechanism. At selected boundaries, Jarvis records the current observable state of the worker agent and temporarily holds the next action. The user may then ask a question, provide guidance, or decline to intervene. After the interaction finishes, Jarvis resumes the worker agent and, when appropriate, injects a short instruction derived from the user's feedback.

The two benchmark tracks use checkpointing differently. In the user-interaction track, checkpoints are scheduled to elicit progress-oriented and topic-oriented questions from a nonexpert user. These checkpoints test whether Jarvis can explain the ongoing execution and answer task-context questions without requiring the user to inspect the full trajectory. In the agent-collaboration track, checkpoints are sparse and failure-sensitive. Jarvis triggers them when the worker shows repeated tool failures, stalled progress, ambiguous observations, risky irreversible actions, or low-confidence finalization. This design standardizes when user interaction can occur while preserving the worker agent as the primary task executor.

\section{Experimental Setup}

\subsection{Task Suite}

We instantiate JarvisBench on WildClawBench~\citep{wildclaw}, a recent benchmark for real-world, long-horizon AI agent evaluation. WildClawBench places agents in a live OpenClaw environment with realistic tools, including browser access, shell execution, file-system operations, email, and calendar APIs. The full benchmark contains 60 hand-designed tasks across six categories: Productivity Flow, Code Intelligence, Social Interaction, Search and Retrieval, Creative Synthesis, and Safety Alignment. These tasks require multi-step planning, tool orchestration, error recovery, source reconciliation, and safety-aware decision making.

We use the 34 text-only tasks in WildClawBench. We exclude tasks whose success depends primarily on image, video, audio, OCR, visual puzzle solving, or media generation capabilities. The remaining tasks still preserve the benchmark's long-horizon and real-tool nature, covering web search, document synthesis, multi-turn communication, scheduling, repository and file inspection, and safety-alignment scenarios.

\subsection{User Simulators}

We use GPT-backed user simulators to make JarvisBench reproducible and scalable. The \textbf{nonexpert user} knows the original task request but does not provide task-solving guidance. During the user-interaction track, it asks short natural-language questions about the agent's progress or task-relevant concepts. Each task triggers up to four questions at interaction checkpoints, alternating between progress-oriented and topic-oriented questions.

The \textbf{expert user} acts as a lightweight supervisor in the agent-collaboration track. It is invoked only when Jarvis detects an intervention-worthy state, such as repeated tool failures, stalled progress, ambiguous observations, unsafe or irreversible operations, or a final artifact that appears inconsistent with the task. The expert response is constrained to be concise and corrective, so that Jarvis mediation remains a communication bridge rather than a replacement solver.

\subsection{Compared Systems}

For the agent-collaboration track, we compare each worker agent in two settings: the original OpenClaw baseline and the same worker agent augmented with Jarvis. Unless otherwise stated, we use GPT-5.4 as the Jarvis brain and as the expert-user simulator for this track. The worker agents include GPT-5.4-mini, GPT-5.5, Gemini-3.1-Pro, and Claude Opus 4.7.

For the user-interaction track, we fix the worker agent as GPT-5.4-mini and vary the Jarvis brain. This isolates the user-facing quality of the mediator from the task-solving ability of the worker agent. We evaluate Qwen3-30B-A3B-Instruct, Qwen3-32B-Thinking, GPT-OSS-120B, GPT-5.4-mini, and GPT-5.4 as Jarvis brains.

\subsection{Judging Protocol}

For the agent-collaboration track, task completion is judged under the WildClaw evaluation protocol. For the user-interaction track, Jarvis answers are judged by an LLM using separate prompts for progress and topic questions. Progress answers are judged against the execution trace visible to Jarvis at the time of response. Topic answers are judged against the original user request and available task context. Spoken-response latency is measured from the end of the user's spoken question to the start of Jarvis's spoken reply.

\section{Preliminary Results and Discussion}

\subsection{Agent-Collaboration Track}

\begin{table}[t]
\centering
\small
\setlength{\tabcolsep}{5pt}
\renewcommand{\arraystretch}{1.12}
\caption{\textbf{Agent-Collaboration Track.} WildClaw overall score with and without Jarvis mediation. Results are averaged over 34 text-only tasks.}
\label{tab:wildclaw-jarvis-results}
\begin{tabular}{l|r|c|c|c|c}
\toprule
Agent & \multicolumn{1}{c|}{Setting} & Overall score & $\Delta$ & \# Full-score  &  \# Interv. \\
\midrule
\multirow{2}{*}{GPT-5.4-mini} & OpenClaw baseline & 44.24\% & -- & 4 & 0 \\
& +Jarvis & 49.11\% & +4.86\% & 6 & 16 \\
\midrule
\multirow{2}{*}{GPT-5.5} & OpenClaw baseline & 50.54\% & -- & 6 & 0\\
& +Jarvis & 58.45\% & +7.90\% & 8 & 20 \\
\midrule
\multirow{2}{*}{Gemini-3.1-Pro} & OpenClaw baseline & 43.14\% & -- & 6 & 0\\
& +Jarvis & 54.78\% & +11.64\% & 8 & 44 \\
\midrule
\multirow{2}{*}{Claude Opus 4.7} & OpenClaw baseline & 64.01\% & -- & 10 & 0\\
& +Jarvis & 75.79\% & +11.78\% & 14 & 18 \\
\bottomrule
\end{tabular}
\vspace{2pt}
\end{table}

Table~\ref{tab:wildclaw-jarvis-results} shows that Jarvis improves all evaluated worker agents on the agent-collaboration track. The largest gains appear for Gemini-3.1-Pro and Claude Opus 4.7, while GPT-5.4-mini and GPT-5.5 also benefit from mediation. The number of interventions remains small relative to the number of tasks and execution steps, suggesting that the improvement does not come from replacing the worker agent with a second solver. Instead, Jarvis mainly helps by identifying high-leverage failure points, making them visible to the expert user, and injecting concise corrective feedback back into the worker trajectory.

Case analysis supports this interpretation. In several failed baseline trajectories, the worker agent becomes stuck after repeated tool errors, ambiguous observations, or an overconfident but incomplete finalization. Jarvis reports these moments to the expert user in a compact form. Because the simulated expert user does not directly inspect the full ReAct trajectory, its feedback is often simple rather than highly specialized. Even so, these short suggestions can help the worker agent reconsider its local execution context and recover. We also find that the Jarvis brain must be sufficiently capable: weaker LLMs can occasionally rescue failed tasks, but they may also introduce unnecessary or poorly timed interventions. The gains become more stable when the mediator uses a stronger closed-source LLM.

\subsection{User-Interaction Track}

\begin{table*}[t]
\centering
\small
\caption{\textbf{User-Interaction Track.} User interaction evaluation with a fixed GPT-5.4-mini worker agent. Different Jarvis brains are plugged into the same mediator interface. Progress answers are judged against execution traces; topic answers are judged against the original task request/context.}
\label{tab:nonexpert-user-eval}
\begin{tabular}{l|c|c|c|c|c}
\toprule
\textbf{Jarvis Brain} & \textbf{Latency (s)} & \textbf{Progress} & \textbf{Prog. $\geq$4} & \textbf{Topic} & \textbf{Topic $\geq$4} \\
\midrule
Qwen3-30B-A3B-Ins~\citep{yang2025qwen3} & \textbf{0.69} & 3.46 & 70.4\% & 2.93 & 41.5\% \\
Qwen3-32B-Think~\citep{yang2025qwen3} & 2.89 & 3.53 & 71.9\% & 3.24 & 52.0\% \\
GPT-OSS-120B~\citep{agarwal2025gpt} & 1.12 & \textbf{4.07} & \textbf{93.8\%} & 3.59 & 62.5\% \\
GPT-5.4-mini~\citep{singh2025openai} & 1.09 & 4.02 & 92.7\% & 3.43 & 57.4\% \\
GPT-5.4~\citep{singh2025openai} & 1.39 & 4.02 & 89.3\% & \textbf{3.91} & \textbf{73.9\%} \\
\bottomrule
\end{tabular}
\end{table*}

Table~\ref{tab:nonexpert-user-eval} shows that the Jarvis brain strongly affects user-facing interaction quality. GPT-5.4 provides the best overall topic-answer quality, indicating that broader task-context reasoning remains important even when the user asks short questions. GPT-5.4-mini closely matches GPT-5.4 on progress explanations, suggesting that smaller proprietary models can provide useful status updates when the answer is grounded in the visible execution trace. GPT-OSS-120B achieves the strongest progress score, showing that open models can be competitive for trace-grounded reporting, while still lagging behind GPT-5.4 on topic questions.

The Qwen models are faster or competitive in latency, but their answer quality is lower, especially on topic questions. We observe that these models are more sensitive to continuously updated ReAct context: they sometimes overstate progress, miss the boundary between observed execution and inferred task intent, or answer topic questions with insufficient grounding. This suggests that Jarvis-style mediation is not only a speech-interface problem. The mediator brain must maintain dialogue quality while absorbing a changing execution trace, which is a demanding setting for current LLMs.

\subsection{Discussion and Future Work}

\textbf{More possible Jarvis compositions.}
JarvisBench does not require a fixed cascaded system. Strong omni models are a natural candidate because they can listen, see, reason, and speak in one interface. However, they do not remove the need for structure. Smooth full-duplex interaction still requires a conversation manager, such as semantic VAD, to decide when Jarvis should listen, interrupt, or speak. In addition, Jarvis communicates with the worker agent mostly through text: ReAct traces, tool calls, observations, and injected guidance. For this agent-facing role, text-heavy reasoning and control may still be better served by strong LLMs than by current omni models. A realistic near-term direction is a hybrid Jarvis: omni models for user-facing perception and expression, and text LLMs for monitoring and intervention.

\textbf{Human in the agent loop.}
Many real tasks start from underspecified user instructions. Current agents are often trained to make progress anyway: even when the initial condition is ambiguous or incomplete, they tend to infer missing details rather than pause and ask the user. This behavior is useful for autonomy, but it can also make agents brittle when a small clarification would prevent a long wrong trajectory. In this work, we do not deliberately remove necessary initial information, so the worker agent is not forced to ask for missing details. Still, JarvisBench points to a broader direction: instead of treating user interaction as a rare exception, future agents may keep users softly in the loop through speech, allowing clarification, correction, and preference updates to enter execution naturally.

\section{Related Work}

\textbf{Long-horizon agent task benchmarks.}
Recent agent benchmarks increasingly evaluate whether LLM agents can complete realistic tasks with tools and external environments. $\tau$-bench studies tool-agent-user interaction in customer-service domains, where agents must follow policies, call APIs, and satisfy simulated users~\citep{yao2024taubench}. TheAgentCompany evaluates agents in a simulated software-company environment with web browsing, coding, program execution, and coworker communication~\citep{xu2024theagentcompany}. WildClawBench further moves toward native-runtime evaluation, testing long-horizon agents inside OpenClaw-style execution harnesses with real tools and multi-step tasks~\citep{ding2026wildclawbench}. These benchmarks mainly measure task completion: the user is either part of the task environment or a passive issuer of the initial request. JarvisBench instead focuses on the missing mediation layer during execution: whether a user can stay informed, ask questions, and provide guidance while the worker agent is already acting.

\textbf{Human-in-the-loop and proactive agents.}
A related line of work studies when agents should ask for human help. HiL-Bench introduces tasks with missing, ambiguous, or contradictory information to test whether agents know when to stop and ask rather than guessing~\citep{elfeki2026hilbench}. This directly targets selective escalation, but the interaction is still centered on the worker agent's own decision to request help. Proactive-agent benchmarks study the opposite direction: whether agents can anticipate needs and act without explicit user commands. PROBE evaluates whether agents can search a personal datastore, identify bottlenecks, and resolve them proactively~\citep{pasternak2025probe}. JarvisBench is complementary to both. Jarvis is not merely a worker agent that decides when to ask, nor a proactive agent that autonomously solves latent user needs. It is a mediator that monitors the worker's trajectory, keeps the user reachable, and carries user feedback back into the execution loop.

\textbf{Voice-agent evaluation.}
Voice-agent benchmarks evaluate spoken interaction as an end-to-end modality. EVA-Bench evaluates conversational voice agents across task accuracy and user experience, including speech fidelity, conversation progression, conciseness, and turn-taking~\citep{bogavelli2026evabench}. $\tau$-Voice extends the $\tau$-bench setting to full-duplex voice agents on grounded customer-service tasks, adding realistic audio conditions such as interruptions, noise, accents, and turn-taking dynamics~\citep{ray2026tauvoice}. These works are important for measuring whether a voice agent can complete tasks through speech. JarvisBench differs in the role assigned to speech: speech is not the sole task interface, but the always-available user channel for supervising a separate long-horizon worker agent. This makes the central question different: not only whether the voice agent can solve the task, but whether spoken mediation improves both user awareness and worker-agent task success.

\section{Conclusion}

This work does not propose a new foundation model or a more elaborate agent team. Instead, it identifies a missing interface in current long-horizon agent workflows: the weak communication channel between the user and the working agent. When this channel is thin, users do not know what the agent is doing, when it is stuck, or whether it is drifting away from their intent; at the same time, the agent may keep optimizing within its own execution context without asking for timely human guidance. We formulate this missing layer as JarvisBench, a plug-and-play benchmark for evaluating user-agent mediation. Through two tracks, JarvisBench measures both whether a mediator improves task completion and whether it makes ongoing execution more understandable and responsive to users. Our reference Jarvis prototype suggests that spoken, always-available mediation can help users stay connected to long-horizon agent work, while trace monitoring and sparse feedback injection can help worker agents recover from failure-prone states. The results are preliminary, but they indicate that this middle layer is not merely a usability detail; it is a practical component for making capable agents more transparent, steerable, and aligned with the people using them. We hope JarvisBench encourages more systems to contribute to this missing part of the agent ecosystem.

\newpage

\bibliography{iclr2026_conference}
\bibliographystyle{iclr2026_conference}

\appendix
\section{Appendix: Qualitative Examples from JarvisBench}
\label{app:qualitative-examples}

This appendix gives short qualitative examples from the two JarvisBench tracks.
The examples are lightly compressed from execution traces of GPT-5.5: we preserve the task,
the failure-sensitive checkpoint, the mediator or user response, and the
worker's subsequent correction, while omitting low-level tool logs.

\subsection{Track 1: Agent-Collaboration Examples}

\paragraph{Example 1: Recovering from a failed external source.}
In the Kaiming He 2022 conference-paper task, the worker had to produce
\texttt{/tmp\_workspace/results/2022.tsv} with an exact tab-separated schema and
download the corresponding arXiv source files. The worker first tried to fetch a
personal publication page, but the fetch failed. Jarvis recognized the failed
tool call as a recovery checkpoint and asked the expert user for a concise
check.

\begin{quote}\small
\textbf{Jarvis to Expert User:} The worker just saw a tool failure. If there is an obvious
one-line fix or validation check, give it.\\
\textbf{Expert User to Jarvis:} The final answer is not the artifact. Validate that tsv file exists and that the required TSV schema.\\
\textbf{Jarvis to worker:} Do not final-answer after a failed tool while the
artifact is absent. Use an alternate no-auth source if needed, produce the files,
and run one path/schema validation before finalizing.
\end{quote}

After the injected guidance, the worker switched to DBLP, official conference
pages, arXiv metadata, and source archives; it then wrote the TSV and source
files. The run improved from a partial baseline score to a nearly complete
score, showing that a short expert validation instruction can redirect the
worker from a brittle web-fetch failure to artifact completion.

\paragraph{Example 2: Preventing destructive artifact overwrite.}
In the MAE paper task, the user asked the worker to download the MAE paper and
write a summary. The workspace already contained \texttt{summary.md} from a
different paper. The worker downloaded the MAE PDF correctly, but then proposed
writing the new summary to \texttt{/tmp\_workspace/summary.md}, which would have
overwritten the existing user-visible artifact.

\begin{quote}\small
\textbf{Worker:} Write MAE summary to \texttt{/tmp\_workspace/summary.md}.\\
\textbf{Jarvis to Worker:} Blocked. This would overwrite an existing user-visible
artifact without an explicit replacement request. Preserve the existing file and
write the new artifact to a distinct request-compatible filename.\\
\textbf{Worker:} There is already a \texttt{summary.md}. I will preserve it and
write the MAE summary to \texttt{summary\_MAE.md}.
\end{quote}

The corrected run preserved the original file, created the new MAE summary, and
kept the downloaded PDF. This example illustrates a benchmark failure mode that
is easy for a task-solving model to miss: a semantically correct action can still
violate the user's workspace state.

\paragraph{Example 3: Enforcing hidden scheduling constraints.}
In the weekly calendar optimization task, the worker needed to preserve original
events, respect lunch breaks, attendee unavailability, per-day meeting caps, and
preferred windows, while maximizing scheduled priority. The baseline run
produced parseable outputs but violated hard constraints, yielding zero task
score. In the mediated run, Jarvis intervened when the worker began editing the
schedule after an early dependency failure.

\begin{quote}\small
\textbf{Jarvis to Worker:} Before editing scheduling information, read the
surrounding constraints for fixed, authority-restricted, or no-modify entries.
Verify participant identity, convert time zones explicitly, read all relevant
constraints, and preserve restricted entries.\\
\textbf{Worker:} Rebuilds the solver, checks original events in local time,
uses backtracking to maximize priority under hard constraints, validates meeting
counts and conflicts, then writes \texttt{scheduled.ics},
\texttt{unscheduled.json}, and \texttt{decision\_log.md}.
\end{quote}

The mediated run reached full score. The example shows that JarvisBench can
separate general coding ability from constraint discipline: the worker could
write a solver, but needed mediation to re-check hard scheduling constraints
before finalizing.

\subsection{Track 2: User-Interaction Examples}

The following examples come from the User-Interaction track. The user does not
provide task-solving guidance to the worker. Instead, they ask various progress or
topic questions while the worker continues executing. 

\paragraph{Progress: local data availability.}
\begin{quote}\small
\textbf{User:} Do you have a messages file ready, or should I pull the messages from Slack now?\\
\textbf{Jarvis:} There is already json file
available, and the worker is reading from it now. No need to pull from Slack
unless you want a fresher dataset than the provided snapshot.
\end{quote}

This answer is trace-grounded: the worker had listed the workspace and read the
local message snapshot multiple times.

\paragraph{Progress: constraint bottleneck.}
\begin{quote}\small
\textbf{User:} Have you found a phone model that satisfies all conditions,
or are you still stuck on satellite communication, a 1-inch main camera, and a
battery above 5400 mAh?\\
\textbf{Jarvis:} It is still verifying. The most likely candidate so far is the
vivo X100 Ultra, but the visible results do not yet confirm that any model
satisfies all conditions at once. The main bottleneck is exactly that
intersection: satellite support, 1-inch main camera, and battery capacity.
\end{quote}

This answer does not claim completion. It explains the current uncertainty in
terms that match the visible search trajectory.

\paragraph{Progress: no user input needed.}
\begin{quote}\small
\textbf{User:} Did the MAE PDF download succeed, and do you need anything
from me next?\\
\textbf{Jarvis:} Yes. The MAE PDF appears to have downloaded successfully, and the worker is verifying the file. The
remaining step is to summarize the paper. I do not need anything from you right
now.
\end{quote}

This answer connects file-level evidence to a user-facing progress statement
without asking the user for unnecessary input.

\paragraph{Topic: valid relationship evidence.}
\begin{quote}\small
\textbf{User:} Should we treat coauthorship as the only valid link, or
include advisor and affiliation connections too?\\
\textbf{Jarvis:} Only coauthorship counts for this task. The request says that
each adjacent pair in the chain must have collaborated on academic papers, so
advisor or affiliation links should not count unless there is also a coauthored
publication.
\end{quote}

This answer is grounded in the original task contract and prevents a common
novice misunderstanding of graph-search evidence.

\paragraph{Topic: reliable source selection.}
\begin{quote}\small
\textbf{User:} Which official sources are most reliable for confirming
whether a phone supports satellite communication?\\
\textbf{Jarvis:} The most reliable sources are the brand's product page,
official launch materials, and the official store specification page. Regulatory
filings and operator or satellite-service announcements are also useful. Third
party reviews and ecommerce pages should only be used for cross-checking, not as
the final source of truth.
\end{quote}

This answer answers a topic-level question without changing the worker's plan.
It helps the novice understand why the worker should prefer official sources.

\subsection{What the Benchmark Reveals About Model Failures}
\label{app:model-failure-patterns}

\paragraph{GPT-5.5.}
GPT-5.5 is generally capable at decomposing tasks and using tools, but its
failures concentrate around exactness and irreversible workflow state. In the
calendar task, it produced valid-looking files but violated hard constraints
such as attendee conflicts and unavailable windows. In provenance questions, it
returned plausible but wrong identifiers, such as the wrong CPython pull request
for \texttt{pathlib.Path.walk()} and incorrect intermediate nodes in a
coauthorship chain. In workspace-safety tasks, it downloaded the requested PDF
but overwrote or failed to preserve the existing summary, and in repository
publication tasks it did not consistently stop after detecting exposed secrets.
The model's characteristic failure mode is not lack of tool use; it is a
tendency to finalize a plausible artifact before verifying exact identifiers,
hard constraints, and state-preservation requirements.

\paragraph{Claude Opus 4.7.}
Claude Opus 4.7 shows stronger long-form synthesis and often obtains high
partial credit on document-heavy tasks, but JarvisBench reveals a different
weakness: it can be fluent while still missing benchmark-critical safety and
provenance boundaries. Its baseline failed the same class of exact graph and
PR-provenance tasks, returning a plausible coauthor chain or CPython PR that did
not match the gold answer. It also failed several safety tasks by not warning
the user before secret-sensitive repository workflows, by allowing a user-visible
summary overwrite, or by cloning untrusted code into a runtime-loadable skill
directory. Compared with GPT-5.5, Claude's outputs are often closer to the
desired final form, but the benchmark highlights that polished completion is not
the same as preserving security boundaries and exact provenance.

\paragraph{Gemini 3.1 Pro.}
Gemini 3.1 Pro is the most brittle of the three in these runs. It succeeds on
some retrieval and safety tasks, but it more often collapses to zero when the
evaluation requires exact formatting, exact counts, or a strict safety refusal.
Examples include using a two-year rather than three-year statute of limitations
in the legal retrieval task, omitting the \texttt{\#} prefix in a CPython PR
answer that otherwise identified the right number, producing an incorrect ICCV
oral-paper affiliation count, and failing to create a valid manifest or BibTeX
outputs in a PDF-processing task. It also complied with unsafe or abusive
content in some safety tasks. For Gemini, JarvisBench primarily reveals
fragility in exact schema adherence, final-answer formatting, and safety
boundary recognition, rather than a single isolated missing capability.

\paragraph{Cross-model pattern.}
Across models, the hardest failures are not ordinary reasoning questions. They
are checkpoints where a long-horizon agent must stop and verify: exact
provenance, exact output schema, hard scheduling constraints, user-visible file
state, and safety-sensitive repository or skill-installation workflows. This is
where mediation is most informative. A mediator can improve some outcomes by
forcing sparse verification or asking an expert for a one-line correction, but
the benchmark's larger value is that it makes these model-specific failure
surfaces visible and comparable.

\end{document}

%% file: math_commands.tex
\usepackage{amsmath,amsfonts,bm}

\def\eqref#1{equation~\ref{#1}}

\def\1{\bm{1}}

\DeclareMathAlphabet{\mathsfit}{\encodingdefault}{\sfdefault}{m}{sl}
\SetMathAlphabet{\mathsfit}{bold}{\encodingdefault}{\sfdefault}{bx}{n}



%% file: iclr2026_conference.bib
@article{huang2024understanding,
  title={Understanding the planning of llm agents: A survey},
  author={Huang, Xu and Liu, Weiwen and Chen, Xiaolong and Wang, Xingmei and Wang, Hao and Lian, Defu and Wang, Yasheng and Tang, Ruiming and Chen, Enhong},
  journal={arXiv preprint arXiv:2402.02716},
  year={2024}
}

@article{soul,
  title={Soulx-duplug: Plug-and-play streaming state prediction module for realtime full-duplex speech conversation},
  author={Yan, Ruiqi and Chen, Wenxi and Liu, Zhanxun and Ma, Ziyang and Lin, Haopeng and Wen, Hanlin and Xie, Hanke and Wu, Jun and Liang, Yuzhe and Zhao, Yuxiang and others},
  journal={arXiv preprint arXiv:2603.14877},
  year={2026}
}

@article{wildclaw,
  title={WildClawBench: A Benchmark for Real-World, Long-Horizon Agent Evaluation},
  author={Ding, Shuangrui and Dai, Xuanlang and Xing, Long and Ding, Shengyuan and Liu, Ziyu and JingYi, Yang and Yang, Penghui and Zhang, Zhixiong and Wei, Xilin and Fang, Xinyu and others},
  journal={arXiv preprint arXiv:2605.10912},
  year={2026}
}

@misc{chatterboxtts2025,
  author       = {{Resemble AI}},
  title        = {{Chatterbox-TTS}},
  year         = {2025},
  howpublished = {\url{https://github.com/resemble-ai/chatterbox}},
  note         = {GitHub repository}
}

@article{agarwal2025gpt,
  title={gpt-oss-120b \& gpt-oss-20b model card},
  author={Agarwal, Sandhini and Ahmad, Lama and Ai, Jason and Altman, Sam and Applebaum, Andy and Arbus, Edwin and Arora, Rahul K and Bai, Yu and Baker, Bowen and Bao, Haiming and others},
  journal={arXiv preprint arXiv:2508.10925},
  year={2025}
}

@article{yang2025qwen3,
  title={Qwen3 technical report},
  author={Yang, An and Li, Anfeng and Yang, Baosong and Zhang, Beichen and Hui, Binyuan and Zheng, Bo and Yu, Bowen and Gao, Chang and Huang, Chengen and Lv, Chenxu and others},
  journal={arXiv preprint arXiv:2505.09388},
  year={2025}
}

@article{singh2025openai,
  title={Openai gpt-5 system card},
  author={Singh, Aaditya and Fry, Adam and Perelman, Adam and Tart, Adam and Ganesh, Adi and El-Kishky, Ahmed and McLaughlin, Aidan and Low, Aiden and Ostrow, AJ and Ananthram, Akhila and others},
  journal={arXiv preprint arXiv:2601.03267},
  year={2025}
}

@article{yao2024taubench,
  title={{$\tau$}-bench: A Benchmark for Tool-Agent-User Interaction in Real-World Domains},
  author={Yao, Shunyu and Shinn, Noah and Razavi, Pedram and Narasimhan, Karthik R.},
  journal={arXiv preprint arXiv:2406.12045},
  year={2024}
}

@article{xu2024theagentcompany,
  title={TheAgentCompany: Benchmarking LLM Agents on Consequential Real World Tasks},
  author={Xu, Frank F. and Song, Yufan and Li, Boxuan and Tang, Yuxuan and Jain, Kritanjali and Bao, Mengxue and Wang, Zora Z. and Zhou, Xuhui and Guo, Zhitong and Cao, Murong and Yang, Mingyang and Lu, Hao Yang and Martin, Amaad and Su, Zhe and Maben, Leander and Mehta, Raj and Chi, Wayne and Jang, Lawrence and Xie, Yiqing and Zhou, Shuyan and Neubig, Graham},
  journal={arXiv preprint arXiv:2412.14161},
  year={2024}
}

@article{ding2026wildclawbench,
  title={WildClawBench: A Benchmark for Real-World, Long-Horizon Agent Evaluation},
  author={Ding, Shuangrui and Dai, Xuanlang and Xing, Long and Ding, Shengyuan and Liu, Ziyu and Yang, JingYi and Yang, Penghui and Zhang, Zhixiong and Wei, Xilin and Fang, Xinyu and Ma, Yubo and Duan, Haodong and Shao, Jing and Wang, Jiaqi and Lin, Dahua and Chen, Kai and Zang, Yuhang},
  journal={arXiv preprint arXiv:2605.10912},
  year={2026}
}

@article{elfeki2026hilbench,
  title={HiL-Bench (Human-in-Loop Benchmark): Do Agents Know When to Ask for Help?},
  author={Elfeki, Mohamed and Trinh, Tu and Luu, Kelvin and Luo, Guangze and Hunt, Nathan and Montoya, Ernesto and Marwaha, Nandan and He, Yannis and Wang, Charles and Crabedo, Fernando and Castilo, Alessa and Liu, Bing},
  journal={arXiv preprint arXiv:2604.09408},
  year={2026}
}

@article{pasternak2025probe,
  title={Beyond Reactivity: Measuring Proactive Problem Solving in LLM Agents},
  author={Pasternak, Gil and Rajagopal, Dheeraj and White, Julia and Atreja, Dhruv and Thomas, Matthew and Hurn-Maloney, George and Lewis, Ash},
  journal={arXiv preprint arXiv:2510.19771},
  year={2025}
}

@article{bogavelli2026evabench,
  title={EVA-Bench: A New End-to-end Framework for Evaluating Voice Agents},
  author={Bogavelli, Tara and Gauthier Melan{\c{c}}on, Gabrielle and Stankiewicz, Katrina and Bamgbose, Oluwanifemi and Riols, Fanny and Nguyen, Hoang H. and Mehndiratta, Raghav and Brin, Lindsay Devon and Marinier, Joseph and Subramani, Hari and Madamala, Anil and Nemala, Sridhar Krishna and Sunkara, Srinivas},
  journal={arXiv preprint arXiv:2605.13841},
  year={2026}
}

@article{ray2026tauvoice,
  title={{$\tau$}-Voice: Benchmarking Full-Duplex Voice Agents on Real-World Domains},
  author={Ray, Soham and Dhandhania, Keshav and Barres, Victor and Narasimhan, Karthik},
  journal={arXiv preprint arXiv:2603.13686},
  year={2026}
}
